\documentclass[letterpaper]{article} % DO NOT CHANGE THIS
\usepackage[]{aaai24}  % DO NOT CHANGE THIS
\usepackage{times}  % DO NOT CHANGE THIS
\usepackage{helvet}  % DO NOT CHANGE THIS
\usepackage{courier}  % DO NOT CHANGE THIS
\usepackage[hyphens]{url}  % DO NOT CHANGE THIS
\usepackage{graphicx} % DO NOT CHANGE THIS
\usepackage{natbib}  % DO NOT CHANGE THIS AND DO NOT ADD ANY OPTIONS TO IT
\usepackage{caption} % DO NOT CHANGE THIS AND DO NOT ADD ANY OPTIONS TO IT
\usepackage{algorithm}
\usepackage{algorithmic}

\usepackage{newfloat}
\usepackage{listings}
\DeclareCaptionStyle{ruled}{labelfont=normalfont,labelsep=colon,strut=off} % DO NOT CHANGE THIS
\floatstyle{ruled}
\newfloat{listing}{tb}{lst}{}
\floatname{listing}{Listing}
\usepackage{subcaption}
\usepackage{booktabs}
\usepackage{multirow}
\newcommand{\nrm}[0]{N\hspace{-0.3em}R\hspace{-0.15em}M}
\newcommand{\rgb}[0]{R\hspace{-0.1em}G\hspace{-0.15em}B}
\newcommand{\xyd}[0]{X\hspace{-0.1em}Y\hspace{-0.15em}D}

\usepackage{amsmath}
\usepackage{amssymb}
\usepackage{xcolor}

\definecolor{darkgreen}{rgb}{0,0.5,0.2}
\definecolor{darkblue}{rgb}{0,0.3,0.6}

\title{Geo6D: Geometric Constraints Learning for 6D Pose Estimation}
\author{
  Jianqiu Chen\textsuperscript{\rm 1},
  Mingshan Sun \textsuperscript{\rm 2},
  Ye Zheng\textsuperscript{\rm 3},
  Tianpeng Bao \textsuperscript{\rm 2},
  Zhenyu He\thanks{Corresponding author.}\textsuperscript{\rm 1},\\
  Donghai Li \textsuperscript{\rm 2},
  Guoqiang Jin \textsuperscript{\rm 2},
  Zhao Rui \textsuperscript{\rm 2},
  Liwei Wu \textsuperscript{\rm 2},
  Xiaoke Jiang \textsuperscript{\rm 4}
}
\affiliations{
    \textsuperscript{\rm 1}Harbin Institute of Technology, Shenzhen \\
    \textsuperscript{\rm 2}SenseTime Research \\
    \textsuperscript{\rm 3}JD.com, Inc \\
    \textsuperscript{\rm 4}International Digital Economy Academy (IDEA)\\}

\usepackage{bibentry}
\begin{document}
\maketitle
\begin{abstract}

% Several 6D pose estimation methods have been proposed to directly regress the object's target pose parameters by building an end-to-end mapping between the input image and output pose.
%
Numerous 6D pose estimation methods have been proposed that employ end-to-end regression to directly estimate the target pose parameters. Since the visible features of objects are implicitly influenced by their poses, the network allows inferring the pose by analyzing the differences in features in the visible region. However, due to the unpredictable and unrestricted range of pose variations, the implicitly learned visible feature-pose constraints are insufficiently covered by the training samples, making the network vulnerable to unseen object poses.
To tackle these challenges, we proposed a novel geometric constraints learning approach called Geo6D for direct regression 6D pose estimation methods. It introduces a pose transformation formula expressed in relative offset representation, which is leveraged as geometric constraints to reconstruct the input and output targets of the network. These reconstructed data enable the network to estimate the pose based on explicit geometric constraints and relative offset representation mitigates the issue of the pose distribution gap.
Extensive experimental results show that when equipped with Geo6D, the direct 6D methods achieve state-of-the-art performance on multiple datasets and demonstrate significant effectiveness, even with only 10\% amount of data.

\end{abstract}
\section{Introduction}

6D pose estimation has drawn widespread attention as the essential prerequisite for emerging applications, such as robotic manipulation, autonomous driving, and augmented reality~\cite{geiger2012we,xu2018pointfusion,chen2017multi}. 
In the computer vision community, several approaches have been proposed to estimate the transformation pose from the object frame to the camera frame. These existing methods can be categorized into two distinct groups: indirect and direct approaches.
Indirect methods~\cite{densefusion,pvn3d,ffb6d} usually first predict an intermediate feature and then use post-processing optimization algorithms, such as least-squares fitting and iterative Perspective-n-Point (PnP) algorithms~\cite{su2022zebrapose, haugaard2022surfemb, li2019cdpn, kiru2019pixel, rad2017bb8}, to calculate the target pose based on the transformation or projection equation. 
The direct methods~\cite{uni6d,li2018deepim,self6d,gdrnet,uni6dv2, es6d} directly predict the final 6D pose parameters (\emph{e.g},  rotation angles, and translation vectors) by the neural network in an end-to-end manner. 
\begin{figure}[!t]
    \centering
    \includegraphics[width=0.49 \textwidth]{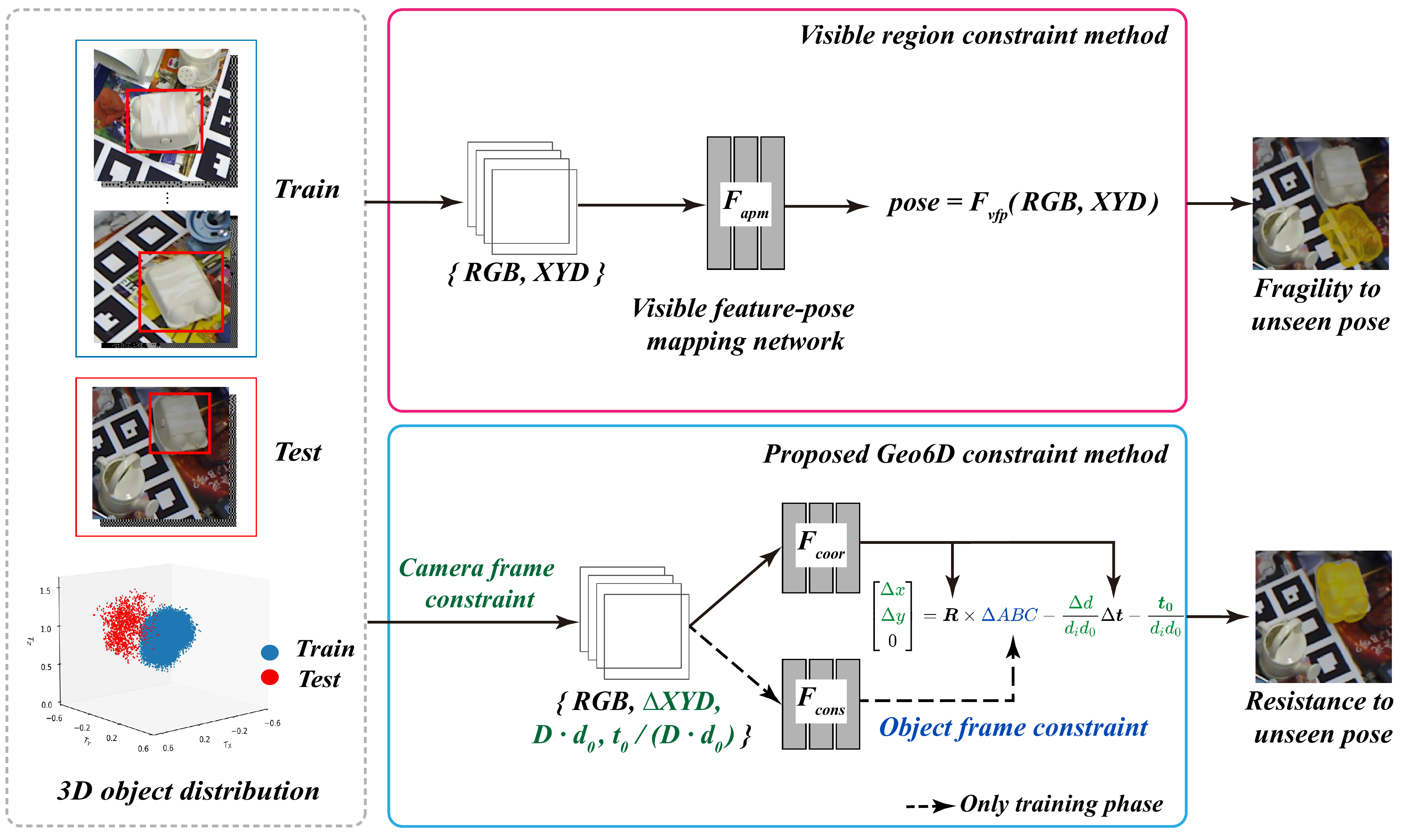}
    \caption{%The proposed geometric constraint-based 6D (Geo6D) pose estimation method. 
    Existing direct pose estimation methods adopt implicit visible feature-pose constraints, that the visible feature of objects depends on its pose, having the fragility to unseen pose. The Geo6D approach introduces 
    novel geometric constraints to rebuild the input and optimization target of the network. It enables the network to regress the pose from explicit geometric constraints and show the resistance to unseen poses.}
    % \caption{The proposed geometric constraint-based 6D (Geo6D) pose estimation method. Different from existing methods of training by the visible feature-pose constraints, the proposed Geo6D approach based on proposed pose transformations introduces geometric constraints to rebuild the network input data (\textcolor{darkgreen}{green part}) as the camera frame constraints and apply a regression head as the object frame constraints (\textcolor{darkblue}{blue part}). The relative offset representation formulate has resistance to the out-of-distribution pose.}

    \label{fig:fig1}
\vspace{-10pt}
\end{figure}
However, both methods have their respective weaknesses.
Although indirect approaches build the geometric constraints by intermediate geometric features, the detached two-stage pipeline makes the optimization target suboptimal, and the time-consuming iterative fitting in the pose estimation stage is an impediment in reality.
On the contrary, the direct methods have the advantage of efficiency and end-to-end optimization.
As shown in Fig~\ref{fig:fig1}, these leverage the implicit constraint to estimate the pose that the visible region of an object in the input image implicitly depends on the object CAD model, the camera intrinsic, and its pose. 
Given that the CAD model and camera intrinsic parameters are commonly known, the network is capable of estimating the pose based on the visible region of the object. 
However, the distribution of poses, specifically the translation component, is unpredictable and unrestricted. The method of mapping visible region appearance features to poses cannot cover every possibility. This creates a vulnerability to differences between the distribution of poses in training and testing data, as well as gaps in appearance domains. This limits the robustness and accuracy of the system.

To improve the visible feature-pose constraints, recent methods~\cite{es6d, pprnet, pprnet++} solve the ambiguity of multiple ground-truth poses relating to the same visible feature by modeling symmetric objects. 
After that, some methods~\cite{uni6dv2, es6d} attempt to leverage an instance segmentation to mitigate the impact of visible features differences arising from camera intrinsic factors, the difference in object pose (camera extrinsic) still hampers the capacity of the model to accurately regress the pose from visible features.
Besides, several direct methods~\cite{self6d, gdrnet} introduce an auxiliary loss to regress intermediate geometric features, such as 2D-3D correspondences, akin to indirect methods. However, these geometric features are not complete constraints to enable the network to regress the pose parameters based on it.
%
% One reason is that the direct method can regress the pose from the visual feature due to the appearance constraints. As shown in the yellow part of Fig~\ref{fig:fig1}, the appearance constraints are based on the same object in different poses having different appearances captured by the camera.
% %
% The neural network is optimized to build a latent appearance-pose mapping function by a large number of training samples covering the potential pose in the test scene.
% However, the translation part in the pose parameter is the same as the object center in the camera frame, whose distribution is unlimited, continuous, and sparse. 
% When there is an object with an out-of-distribution pose in the test scene, the mapping function is missing, leading to the limitation of robustness and precision. %

To solve these issues, we proposed the geometric constraints (Geo6D) learning approach that introduces a reformulated pose transformation to establish robust constraints on both camera and object frames by a relative offset representation.
Specifically, the proposed Geo6D constraints are built upon the pose transformation formula.  The rigid object points' 3D coordinates on the different frames can be transformed based on the pose. 
To address the distribution gap, we introduce a reference point and reformulate the pose transformation formula from the camera frame 3D coordinate representation (the offset for the visible point to the camera) to a relative offset for the visible point to the selected reference point.
% and appending the variables which can be calculated from the input camera frame object coordinates as the additional inputs. 
For making the formula learning-friendly and mathematically correct during network fitting, we separate the variables based on coordinate frames as explicit geometric constraints, demonstrated in Fig~\ref{fig:fig1}. 
For the camera frame variables, we supply and linearize all required variables in the camera frame as input and feed them to the network.
For the object frame constraints, we introduce an additional regression network output head to predict the corresponding relative offset value in the object frame.

% The relative representation addresses the pose distribution problem. Different from coordinate normalization, the Geo6D constrict reformulates the input and output data enabling it to follow the pose transformation formula in relative representation.  \note{formula} Besides, the additional camera frame constraints input and camera frame constraints optimization make it learning-friendly and mathematically correct. 

We encapsulate the Geo6D mechanism as a plugin, which rebuilds the input and output targets of the network and integrates it with two pose estimation networks.
Extensive experiments demonstrate the effectiveness of our method, without sacrificing efficiency in both training and inference to enhance accuracy and stability and reduce the required amount of training data.
%
% Besides the theoretical contribution, we plugin the explicitly geometric constraints into two pose estimation networks, our Geo6D accelerates model convergence and reduces data dependencies. 
%
It only requires 10\% of training data to reach the comparable performance of full training data. Furthermore, we analyze the impact of the Geo6D mechanism from the perspective of the loss function.

To summarize, our main contributions are:
% \begin{itemize}
%     \item Propose a geometric constraint learning approach to establish explicit geometric constraints between the input and the regression target for direct methods.
%     \item Introduce additional inputs and optimization targets to make the network learning-friendly and mathematically correct.
%     \item Extensive experimental results demonstrate that the proposed Geo6D effectively improves existing direct regression methods' performance to achieve state-of-the-art overall results and reduce the required training data, making it more practical for real-world applications.
% \end{itemize}

\begin{itemize}
    \item Introducing a pose transformation formula in a relative offset presentation to establish explicit geometric constraints for direct methods.
    \item Proposing the Geo6D mechanism, a plugin module that processes input data and optimization targets to adhere to the geometric constraints, making the network learning-friendly and mathematically correct. 
    \item Extensive experimental results demonstrate that the proposed Geo6D effectively improves the accuracy of existing direct pose estimation methods achieving state-of-the-art overall results and reducing the training data requirement, thus making it more practical for real-world applications.
\end{itemize}
\section{Related work}
% We review the indirect and direct 6D pose estimation methods and the current geometric constraints in this section.

\subsection{Indirect 6D pose estimation} 
Indirect methods first predict intermediate geometric information and then exploit the projection constraints to estimate the 6D pose by optimization function. 
Recent methods~\cite{pvnet,pvn3d, ffb6d}  introduce the keypoints mechanism in 6D pose estimation and then estimate the 6D pose by a least-squares fitting algorithm, which takes advantage of the geometric constraints of rigid objects to train the keypoint prediction network. 
Different from the keypoints-based methods, 2D-3D correspondence-based methods~\cite{su2022zebrapose, hodan2020epos, haugaard2022surfemb, li2019cdpn, kiru2019pixel, rad2017bb8} first establish the correspondences between 2D coordinates in the image plane and 3D coordinates in the object coordinate system by the neural network and then solve the 6D pose by a PnP or RANSAC algorithm. 
However, these indirect methods are only optimized in the first stage rather than the final pose regression, which is suboptimal compared with direct methods. Moreover, the optimization is time-consuming and computationally expensive in practical applications.

\subsection{Direct 6D pose estimation} 
To estimate 6D pose efficiently, recent approaches~\cite{es6d,uni6d,li2018deepim,self6d,gdrnet} directly regress the final 6D pose parameters from the neural network instead of intermediate results. 
Densefusion~\cite{densefusion} extracts the visible region features information from RGB-D images by two separate backbones to extract the features from 2D and 3D spaces and fuses them with a dense fusion network. 
Uni6D~\cite{uni6d} simplifies the architecture with a homogeneous single backbone to process RGB-D data, by introducing the extra UV data into input to preserve the projection constraints.
Since the corresponding visible features of the object and pose are sensitive to the visual ambiguity of the symmetric object, there are multiple ground-truth poses related to the same visible features that confused the network fitting. ES6D individually models different types of symmetric objects to solve the issue of multiple pose mapping to the same visible features.
Besides, the camera intrinsic is another factor for visible features of the object, Uni6Dv2 adopts an instance segmentation method to mitigate the impact of visible features difference from the camera intrinsic difference.
However, since the pose parameters are unpredictable and unrestricted and the visible features to pose mapping can not be exhaustive and fragile for the unseen pose of the object in the test scene.
To enhance network training, some methods ~\cite{self6d, gdrnet} leverage the intermediate geometric features, i.e. 2D-3D correspondences, akin to indirect methods as an auxiliary task to help network fit the pose transformation. However, the constraints from intermediate geometric features are insufficient to enable the network to regress the pose parameters based on it. 
%
% To alleviate the aforementioned problem, ES6D~\cite{es6d} densely regresses offsets from visible points to the centroid point in 3D space. However, The translation part was eliminated in the calculation of the offset, making it an ill-conditioned problem and affecting the convergence of the regression network. 

\subsection{Geometric constraints in 6D pose estimation}
The indirect methods usually utilize different geometric constraints to train a neural network to predict intermediate features such as predefined keypoints and 2D-3D correspondence. It has an explicit correlation between the input and output target features, which is easier for network optimization. Unlike the indirect method, the direct method needs to regress the pose parameters by the network, which restricts the network output target. To provide more geometric constraints information, some methods~\cite{self6d,gdrnet} apply an additional network output head and auxiliary loss to regress intermediate geometric features such as 2D-3D correspondences matrix or render alignment.
However, the geometric correlations between the network output pose parameters and input data still are not explicitly established.
%
% Hence, we propose the Geo6D mechanism reformulating the input and output to hold the geometric constraints in the relative offset representation and supplement additional geometric values into input to make it easy to be fitted by the network.
Hence, we propose the Geo6D mechanism to establish an explicit geometric constraint by reformulating the input and output to hold the proposed relative offset representation pose transformation formula and make the pose estimation process easily trainable by the network.

\section{Method}
% \begin{figure}[t]
%     \centering
%     \includegraphics[width=0.4 \textwidth]{figures/5d_anchor.png}
%     \caption{The proposed  reference point in 2D-3D projection process. \textcolor{red}{Red dots} denote a  reference point and \textcolor{blue}{Blue dots} denote a visible point.}
%     \label{fig:anchor_mechanism}
% \end{figure}
The Geo6D mechanism is proposed to enable the network to regress the pose in an end-to-end manner with explicit geometric constraints.
In this section, we first introduce our geometric constraint-based 6D pose estimation method. Then we present how to adopt the proposed constraints in these direct methods, and finally, we analyze the impact of the Geo6D mechanism from the perspective of the loss function that balances of rotation and translation part training losses. 
\begin{figure*}[tbp]
    \centering
    \includegraphics[width=0.9 \textwidth]{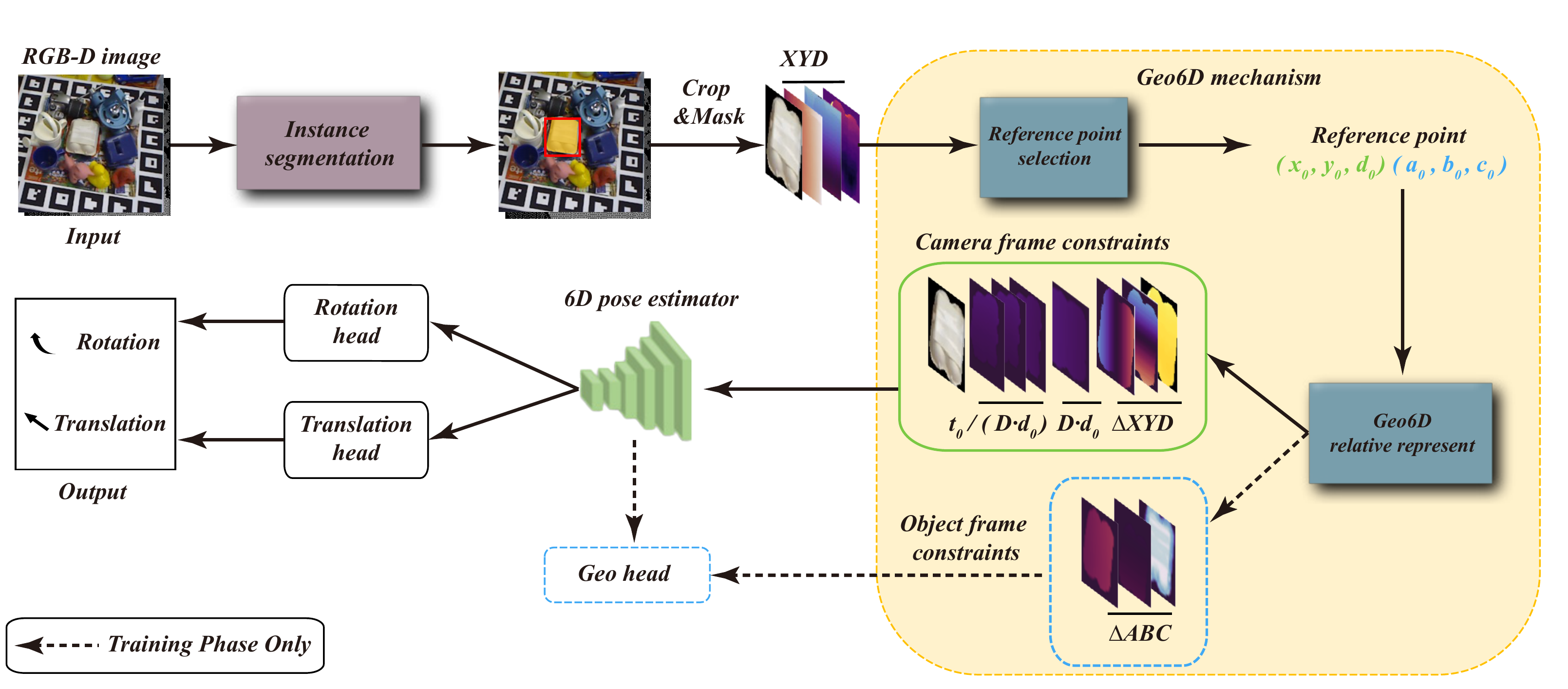}
    \vspace{-5pt}
    \caption{Overview of our proposed Geo6D method. The yellow area depicts the Geo6D mechanism, which reformulates the network input and output target based on Eq~(\ref{equ:new_prj}). 
    The proposed Geo6D constraints are divided into two parts (the camera frame and object frame constraints) according to whether the introduced variables can be captured during the inference phase.
    %: the camera frame constraints which are fed into the network and the object frame constraints which are regressed by an additional network head.
    %It is divided by the coordinate frame, where the camera frame constraints are fed into the network and the object frame constraints are optimized by an additional network output head.
    }
    \label{fig:pipline}
    \vspace{-10pt}
\end{figure*}

%\subsection{Revisit the direct pose estimation methods}
\subsection{Geometric constraints for direct pose estimation}
Current end-to-end direct regression methods~\cite{uni6dv2, es6d} usually adopt a two-stage pipeline. 
In the first stage, an instance segmentation is utilized to crop and mask the candidate objects from the RGB-D image. 
In the second stage, the 3D coordinates $\xyd$ of each object, projected from the depth image and concatenated with the $\rgb$ channel are fed into the pose regression network to directly regress the pose parameters, i.e. the rotation matrix $\boldsymbol{R} \in SO(3)$ and translation vector $\boldsymbol{t} \in \mathbb{R}^3$.
%After that, based on the pin-hole project equation calculate the 3D coordinate in the camera frame from the depth image. The object's 3D coordinates \xyd are then extracted and concatenated with the RGB channel taken as the input of the pose regression network. The network output is the pose parameters, e.g. rotation matrix ($\boldsymbol{R}$) and translation vector ($\boldsymbol{t}$), and leverage the transformation loss, e.g. ADD-Loss to optimize the pose parameters. 

According to the pose transformation formula
\begin{equation}\begin{aligned}\label{equ:translation_equ}
    \begin{bmatrix}
         x \\
         y \\
         d
    \end{bmatrix}
    =  \boldsymbol{R} \times \begin{bmatrix}
      {a}\\
      {b}\\
      {c}
    \end{bmatrix}+{\boldsymbol{t}},
\end{aligned}\end{equation}
any visible point on the target object surface with the camera frame coordinates $(x_i,y_i,d_i)$  has its corresponding object frame coordinates $(a_i,b_i,c_i)$.
%
%Similarly, we select the mean values of these visible point coordinates in the camera frame as the reference point $(x_0,y_0,d_0)$ and based on Eq~(\ref{equ:translation_equ}) the object frame coordinate for selected reference point $(a_0,b_0,c_0)$ can be calculated by the ground truth $\boldsymbol{R}$, $\boldsymbol{t}$ during training phase.
Similarly, the camera frame coordinate $(x_0,y_0,d_0)$ and the corresponding object frame coordinate  $(a_0,b_0,c_0)$ of the centroid of all visible points should satisfy Eq~(\ref{equ:translation_equ}).
We can obtain the following naive geometric constraints by substituting coordinates of any visible point $(x_i,y_i,d_i)$ and the reference point $(x_0,y_0,d_0)$ into Eq~(\ref{equ:translation_equ}) successively and then subtracting both sides equally:
\begin{equation}\begin{aligned}\label{equ:translation_equ2}
    \begin{bmatrix}
        x_i - x_0 \\
        y_i - y_0 \\
        d_i - d_0
    \end{bmatrix}
    = \boldsymbol{R} \times \begin{bmatrix}
      {a_i - a_0}\\
      {b_i - b_0}\\
      {c_i - c_0}
    \end{bmatrix}. % + (\boldsymbol{t} - \boldsymbol{t}).
\end{aligned}\end{equation}
%However, directly adopting the relative offset value between the visible point and the reference point will lead to the translation part being eliminated and excluded from the optimization. To avoid this issue, we select the norm frame coordinates which simultaneously divide both sides of the equation by their depth value $d_i$ and $d_0$. Then, the pose transformation formula can be reformulated as the following equation by simultaneously subtracting the reference point coordinates on both sides of the equation/
However, these constraints only preserve the rotation part of the pose parameters and eliminate the translation part, which provides no benefit to regressing the $\boldsymbol{t}$ during training.
To preserve constraints of the rotation and translation at the same time, we scale all coordinates with respect to their corresponding depth value $d_i$ or $d_0$ before subtracting and resulting in the following geometric constraints:
\begin{equation}\begin{aligned}\label{equ:new_prj}
    \begin{bmatrix}
        \Delta x \\
        \Delta y \\
        0
    \end{bmatrix}
    = \boldsymbol{R} \times \Delta ABC -\frac{\Delta d}{d_{i}d_{0}} \Delta \boldsymbol{t} -\frac{\boldsymbol{t_0}}{d_{i}d_{0}}
\end{aligned}\end{equation}
where
\begin{equation*}
\Delta x=\frac{x_i}{d_i}-\frac{x_0}{d_0},~~\Delta y=\frac{y_i}{d_i}-\frac{y_0}{d_0},~~\Delta d=d_i-d_0,
\end{equation*}
\begin{equation*}
  \Delta ABC=\begin{bmatrix}
     \frac{a_{i}}{d_{i}}-\frac{a_{0}}{d_{0}} \\
     \frac{b_{i}}{d_{i}}-\frac{b_{0}}{d_{0}} \\
     \frac{c_{i}}{d_{i}}-\frac{c_{0}}{d_{0}}
    \end{bmatrix},
~~\Delta \boldsymbol{t}=\boldsymbol{t}-\boldsymbol{t}_0, 
~~\boldsymbol{t}_0=\begin{bmatrix} x_0 \\ y_0 \\ d_0 \end{bmatrix}.
\end{equation*}
%
%It holds the geometric constraints of pose estimation in the relative representation. 
%For making the formula learning-friendly and mathematically correct during network fitting, 
As shown in Fig~\ref{fig:pipline}, we divide the variables in Eq~(\ref{equ:new_prj}) into two groups according to whether they can be captured during the inference phase. 
Since $\Delta x$, $\Delta y$, $\Delta d$, $d_id_0$ and $\frac{\boldsymbol{t_0}}{d_i \cdot d_0}$ can be calculated from the camera frame coordinates, these variables can be fed into the network directly as the \textbf{camera frame constraints}. 
On the contrary, the object frame coordinates $a_i, b_i, c_i$ can not be captured during the inference phase and these variables are regressed by the auxiliary network head as the \textbf{object frame constraints}. 
As illustrated in Fig~\ref{fig:pipline}, the object frame constraints and the Geo Head are activated during the training phase but deactivated during inference.
The camera frame constraints and object frame constraints together make up the proposed geometric constraints.
%Since the object frame coordinates $a_i, b_i, c_i$ are corresponding with the camera frame coordinates by the ground truth pose parameters, the object frame constraints can not be captured during the test phase. Hence, the $\Delta ABC$ is taken as an additional optimization target estimated by an additional network head as shown in Fig~\ref{fig:pipline}. It not only provides the object frame constraint to the network by optimization but also can be removed during inference. The camera and object frame constraints consist of the proposed geometric constraints.
Because the proposed Geo6D mechanism just modifies the input and output without any assumption of the network architecture, it can be plugged into 6D pose direct regression methods described in the following section.
% Further comparisons of our Geo6D mechanism design are presented in our ablation study. 
%
% To establish explicit geometric constraints, we adopt the $\Delta ABC$ serving as an additional regression target of Geo head to optimize and learn the transformation equation.
% Taking into account the missing variables $d_id_0$ and $\frac{t_0}{d_id_0}$ in the input data, we introduce them as additional inputs to linearize and simplify the optimization function, making it easier to fit the network. 

\subsection{Learning framework}
In this section, we show how to integrate the Geo6D mechanism into current RGB-D direct regression methods.
The overall framework is depicted in Fig~\ref{fig:pipline}, with our proposed Geo6D mechanism in yellow.
Different from the current methods taking the $\rgb$  and $\xyd$  as the network input, the Geo6D mechanism is based on the geometric constraints to adopt the camera frame constraint variables $\Delta \xyd, D \cdot d_0, \frac{\boldsymbol{t_0}}{D \cdot d_0}$ concatenated with $\rgb$ as input, where $D$ and $\Delta \xyd$ stands for the set of all points' depth value $d_i$ and the offset value $\Delta x, \Delta y, \Delta d$.
An additional output head (Geo head) introduces object frame constraints in order to regress the $\Delta ABC$.
The Geo head is made up of convolutional blocks that are used to regress the visible points' relative offset in the object frame and optimize using an L2 loss. To establish explicit geometric restrictions, the object frame constraints collaborate with the camera frame constraints in the input data. It explicitly presents the necessary variables as input or optimization targets, allowing the network to regress the pose parameters $\boldsymbol{R}$ and $\boldsymbol{t}$ from them.
Furthermore, the pose estimator output changes the translation vector $\boldsymbol{t}$ with $\ominus \boldsymbol{t}$, which the final translation vector $\boldsymbol{t}$ can be deduced by adding the reference point coordinates $\boldsymbol{t_0}$ in the camera frame. 
%The pose estimation network input and output data can be expressed using the suggested Geo6D technique as follows:
The pose estimation network can be summarised as the following function after utilizing the proposed Geo6D mechanism:
\begin{equation} \label{equ:new_maping}
\boldsymbol{R},\ominus \boldsymbol{t}, \Delta ABC =f(\rgb,\Delta \xyd, D \cdot d_0, \frac{\boldsymbol{t_0}}{D \cdot d_0} ).
\end{equation}

\subsection{Balanced ADD loss}
For most approaches, ADD loss is used to train the network to predict 6D pose. It transforms the points sampled on the object CAD model by the predicted pose and ground truth pose in a heterogeneous manner, and minimizes the distances between matching points in two separate transformation point sets as follows:
\begin{figure}[t]
    \centering
    \includegraphics[width=0.45 \textwidth]{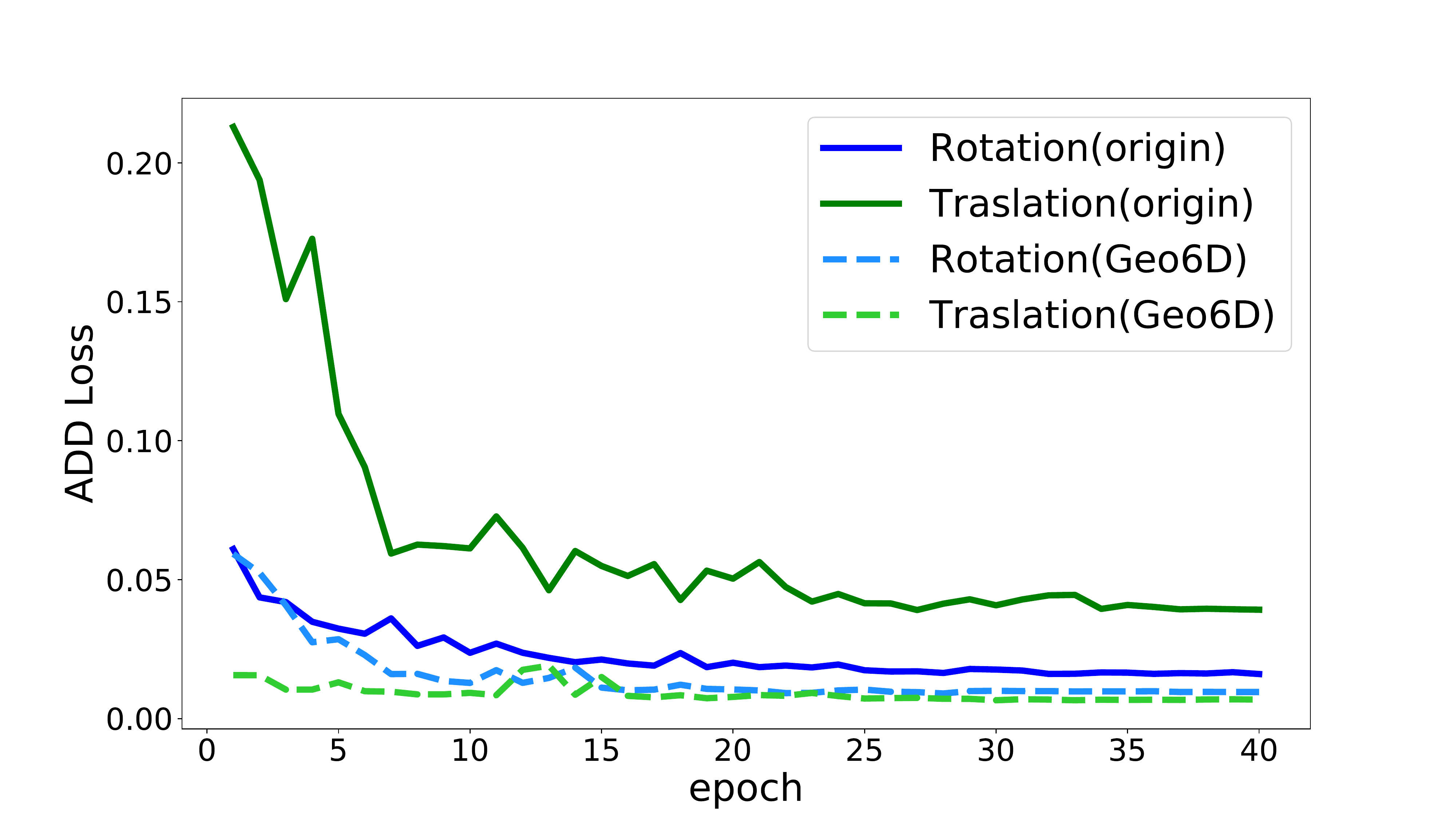}
    \vspace{-5pt}
    \caption{The Geo6D balances ADD loss of the rotation part and translation part on the Occlusion LineMOD dataset.}
    \vspace{-10pt}
    \label{fig:balanced_loss}
\end{figure}
\begin{equation}\label{equ:add_loss}
\begin{split}
L_{ADD}=\frac{1}{m} \sum_j\|\left(\boldsymbol{\hat{R}} p_j+\boldsymbol{\hat{t}}\right)-(\boldsymbol{\bar{R}}p_j+\boldsymbol{\bar{t}})\|^2_2
\end{split}
\end{equation}
where $p_j$ is the $j^{th}$ point from $m$ randomly sampled CAD model's 3D points in the object frame, $(\boldsymbol{\hat{R}}, \boldsymbol{\hat{t})}$ is the predicted pose and $(\boldsymbol{\bar{R}}, \boldsymbol{\bar{t}})$ is the ground truth pose.
Recombining Eq~(\ref{equ:add_loss}), it can be reformulated as: 
\begin{equation}\label{equ:add_loss_recombine}
\begin{split}
L_{ADD} &=\frac{1}{m} \sum_j\|(\ominus\boldsymbol{R} p_j+ \ominus \boldsymbol{t})\|^2_2 \\
 &= \frac{1}{m} \sum_j(\|\ominus\boldsymbol{R} p_j\|^2_2 + 2 \ominus\boldsymbol{R} p_j \ominus \boldsymbol{t} + \|\ominus \boldsymbol{t}\|^2_2) \\
  &= \frac{1}{m} \sum_j(\|\ominus\boldsymbol{R} p_j\|^2_2) + \frac{2}{m}\ominus\boldsymbol{R} \sum_j p_j \ominus \boldsymbol{t} + \|\ominus \boldsymbol{t}\|^2_2
\end{split}
\end{equation}
where 
$\ominus\boldsymbol{R} = \boldsymbol{\hat{R}} - \boldsymbol{\bar{R}},
~~\ominus \boldsymbol{t} = \boldsymbol{\hat{t}} - \boldsymbol{\bar{t}}.
$
Since $p_j$ is sampled from the object frame, the centroid of sampled points is approx to the origin that $\sum_j p_j \approx 0$. 
Hence, the ADD loss is approximately equal to 
\begin{equation}\label{equ:new_add_loss}
\begin{split}
L_{ADD} \approx \frac{1}{m} \sum_j(\|\ominus\boldsymbol{R} p_j\|^2_2) + \|\ominus \boldsymbol{t}\|^2_2
\end{split}
\end{equation}
which consists of the rotation part $\frac{1}{m} \sum_j(\|\ominus\boldsymbol{R} p_j\|^2_2)$ and translation part $\|\ominus \boldsymbol{t}\|^2_2$.

Because any point on the object is still inside the circumscribed sphere of the object after rotating, the rotation component is constrained by the object's diameter $d$.
The translation portion, on the other hand, is unbounded since $\boldsymbol{\hat{t}}$ and$\boldsymbol{\bar{t}}$ are absolute 3D translate vectors, signifying the centroid of the object in the camera frame, which can be any place in 3D space. 
As a result, when $\|\ominus \boldsymbol{t}\|^2_2 \gg d$, 
%As a result, when $\|\ominus \boldsymbol{t}\|^2_2 \gg 2*r$, 
%where $r$ is the radius of the object circumscribing sphere,
the translation part may dominate the optimization while the rotation part is ignored. 
To accommodate the absolute magnitude of the translation vector, the loss will be downgraded to the $l_2$ loss. The ADD loss has an apparent imbalance between the rotation and translation parts, as seen in Fig~\ref{fig:balanced_loss}, with the translation part occupying the majority. 

After adopting the Geo6D, the regression target translation vector is replaced by an equivalent offset value between the object centroid and the reference point, which range is transformed into the range $(-d/2, d/2)$ comparable to the rotation part.
As a result, the rotation and translation parts are balanced in the same magnitude, allowing the rotation part to converge faster and the balanced loss to produce better optimization results. 
%which helps the network pay equal attention to both rotation and translation transformations. 
%
The balanced ADD loss optimizes the regression network to find correspondences between input data, the CAD model rotated, and the offset translate vector.
Hence, the Geo6D mechanism corrects the ADD loss degradation and simplifies the pose estimation task.

\section{Experiments}
\subsection{Benchmark datasets}
\label{sec:datasets}
We conduct our experiments on three benchmark datasets. 

\textbf{LineMOD}~\cite{hinterstoisser2011multimodal} contains 13 sequences of 13 low-textured objects. Since there is only about 1.2k real training data annotated with 6D pose, we follow previous work~\cite{pvnet,posecnn,pvn3d,uni6d} to add synthetic images for training with 99.71\% synthetic ratio.

\textbf{Occlusion LineMOD}~\cite{brachmann2014learning} consists of 1214 testing images 
that are selected from the LineMOD dataset in the occlusion scene. Since there is no extra real training data, the training dataset is the same as LineMOD. The domain gap between training and testing datasets, as well as the heavily occluded objects, make it more challenging for 6D pose estimation.

\textbf{YCB-Video}~\cite{calli2015ycb} is a large and challenging dataset that contains 21 objects with 92 RGB-D sequences. It provides a large amount of real training data holding the same pose distribution between training and testing datasets. 

%
% It exists significant location distribution and style gaps between training and testing datasets, which is challenging for methods to handle these domain gaps.

\subsection{Evaluation metrics}

We adopt the commonly used average distance metrics ADD, ADD-S, and ADD(S) to evaluate different methods.
ADD evaluates the mean of pairwise distance between two object point clouds which are transformed according to the ground truth pose $[\bar{\boldsymbol{R}}, \bar{\boldsymbol{t}}]$ and the predicted pose $[\boldsymbol{\hat{R}}, \boldsymbol{\hat{t}}]$ respectively:
\begin{equation}
    \begin{split}
        \textrm{ADD} = \frac{1}{m}\sum_{p\in\mathcal{O}}||(\boldsymbol{\hat{R}}p +\boldsymbol{\hat{t}}) - (\boldsymbol{\bar{R}}p+\boldsymbol{\bar{t}})||
    \end{split}
\end{equation}
where $\mathcal{O}$ denotes the 3D model of a object, $p$ denotes any point on the model and $m$ denotes total point number in the model.
To alleviate the ambiguous matching of the symmetric objects, ADD-S is adopted to estimate the closest point distance between two point clouds: 
\begin{equation}
    \begin{split}
        \textrm{ADD-S} = \frac{1}{m}\sum_{p_{1}\in\mathcal{O}}\min_{p_{2}\in\mathcal{O}}||(\boldsymbol{\hat{R}}p_{1}+\boldsymbol{\hat{t}}) - (\boldsymbol{\bar{R}}p_{2}+\boldsymbol{{\bar{t}}})||.
    \end{split}
\end{equation}
For convenience, we introduce ADD(S) metric:
\begin{eqnarray}
\textrm{ADD(S)} =
\begin{cases}
\textrm{ADD}   & \mathcal{O}~is~asymmetric\\
\textrm{ADD-S} & \mathcal{O}~is~symmetric 
\end{cases}
\end{eqnarray}
For LineMOD and Occlusion LineMOD datasets, we choose the threshold with 0.1d (10\% of the diameter of the object)  to calculate the accuracy of ADD(S), following~\cite{pvnet,ffb6d}.
%We choose the accuracy of ADD(S) with a maximum threshold of 10\% of the objects' diameter
For the YCB-Video dataset, following~\cite{posecnn,densefusion,pvn3d,ffb6d, es6d}, we calculate the AUC (area under the accuracy-threshold curve) of ADD(S) with a maximum threshold of 0.1 meters.

\subsection{Apply Geo6D to different methods}
To verify the extensibility of the proposed Geo6D, we apply it to ES6D~\cite{es6d} and Uni6Dv2~\cite{uni6dv2}. These two methods have different frameworks, output formats, and loss functions.

\textbf{Apply Geo6D mechanism to Uni6Dv2.}
The input of Uni6Dv2 consists of an image patch $(\rgb, X, Y, D,\nrm)$ which is cropped by the prediction of the segmentation network in the first stage. 
The sparse regression network based on the image patch then regresses the rotation and 3D location of each object in the camera frame and trains it using ADD Loss.
To apply the Geo6D mechanism to Uni6Dv2, we replace the visible points' absolute positional encoding part $(X, Y, D)$ with relative values $(\Delta X,\Delta Y,\Delta D)$ and follow Eq~(\ref{equ:new_prj}) supplements the additional component concatenating with input data among channels.
For the output part, we use an additional Geo head with Conv blocks to regress visible points' $\Delta ABC$ and adjust the regression target of the original translation head reformulated as the offset of the reference point to the centroid $\Delta t$.
The comparison results are shown in Tab~\ref{tab:3data-ADD}.
%
% After adopting the Geo6D mechanism to Uni6Dv2, it shows an obvious improvement(+39.2\%) on the Occlusion LineMOD.
% Due to the advance of network convergence, adopting Geo6D shows significantly greater improvements in fewer amounts of training data demonstrated in Tab~\ref{tab:YCB-AUC-es6d}.
%

\textbf{Apply Geo6D mechanism to ES6D.}
The original ES6D uses a dense regression network to estimate the offsets of visible points to an object's centroid and a confidence score by feeding input cropped RGB images with vanilla normalized $(x,y,d)$. To reach the final pose, it substitutes a loss for the ADD loss and chooses the point with the highest confidence score. We alter the input to be Eq~(\ref{equ:new_prj}) and add an additional Geo head with the same structure as the translation head to regress $\Delta ABC$ in order to apply the Geo6D method to ES6D. Because ES6D does not give the Occlusion LineMOD result, we implement it on the YCB-Video dataset, and all results are based on Ground Truth segmentation masks due to its sensitivity to mask results. We provide the results about the different amounts of training data in Tab~\ref{tab:YCB-AUC-es6d}. 
% Geo6D mechanism leverages explicit geometric constraints and solves the translate part optimization issue in vanilla normalization. 
% Experiments on Tab~\ref{tab:YCB-AUC-es6d} show significantly further improvements in fewer amounts of training data, demonstrating their effectiveness on the dense regression method.
\begin{table}[t]
\begin{center}
\resizebox{\linewidth}{!}{
\begin{tabular}{l|c|c|c|c}
    \toprule
 & Geo6D & Occlusion LineMOD  & LineMOD & YCB-Video  \\
    \midrule
      Uni6Dv2 & w/o & 40.6   &  97.2 & 91.5\\ % & 76.4 \\
      Uni6Dv2 & w & \textbf{79.8}(+39.2)  &  \textbf{99.6}(+2.4) & \textbf{91.6}(+0.1) \\ %  & \textbf{90.1} \\
  \bottomrule
  \end{tabular}}
\end{center}      
\caption{Evaluation results of Uni6Dv2 with Geo6D on Occlusion LineMOD, LineMOD and YCB-Video datasets.}
\label{tab:3data-ADD}
\end{table}

\begin{table}[t]
\begin{center}
\resizebox{0.99\linewidth}{!}{
\begin{tabular}{l|c|c|c|c}
    \toprule
  Setting & Geo6D & 10\% R+10\%S & 10\% R+100\%S  & 100\% R+100\%S \\
      \hline
        ES6D & w/o & 83.6  & 90.1  & 93.2\\
        ES6D & w & \textbf{84.7}(+1.1) & \textbf{92.3}(+2.2) & \textbf{93.6}(+0.4)\\
        \midrule
        Uni6Dv2 & w/o & 79.7  & 88.0  & 91.5 \\
        Uni6Dv2 & w & \textbf{86.5}(+6.8) & \textbf{89.3}(+1.3) & \textbf{91.6}(+0.1) \\
        \bottomrule
\end{tabular}}
\end{center}
\caption{
Evaluation results of ES6D and Uni6Dv2 with Geo6D on YCB-Video dataset with different amounts of training data. R means real data and S means synthetic data.  
}
\label{tab:YCB-AUC-es6d}
\vspace{-0pt}
\end{table}

\subsection{Comparison with SOTA methods}
We provide comprehensive and detailed comparison results on Occlusion LineMOD, LineMOD, and YCB-Video datasets. For the brevity of the paper, category-level experimental results on YCB-Video are provided in Supplementary Materials.

\textbf{Evaluation on Occlusion LineMOD dataset.}
% 这个地方最好和method说的能力能呼应上
The quantitative results on the Occlusion LineMOD dataset are presented in Tab~\ref{tab:ocLineMOD-ACC}. Compared with the baseline Uni6Dv2, employing the Geo6D mechanism achieves a significant improvement (+39.2\%) on ADD(S) metric and outperforms state-of-the-art method FFB6D~\cite{ffb6d}(indirect method) by 13.6\%. Noteworthy, adopting the Geo6D mechanism makes the network more reliable when there are significant translation distribution gaps. 

\textbf{Evaluation on LineMOD dataset.}
The quantitative results on the LineMOD dataset are presented in Tab~\ref{tab:LineMOD-ADD}. Compared with our baseline Uni6Dv2, introducing the Geo6D mechanism brings a 2.5\% performance gain on ADD(S), demonstrating its effectiveness. Compared to FFB6D~\cite{ffb6d}, the performance gap is minimal (99.6\% vs. 99.7\%), while our Geo6D-based direct method has a more straightforward design and faster speed.

% \begin{table*}[tbp]
% \begin{center}
% \resizebox{0.85\linewidth}{!}{
% \begin{tabular}{l|c|c|c|c|c|c|c|c|c}
%     \toprule
%   & PoseCNN & DenseFusion & PVN3D & FFB6D & Uni6D & Uni6Dv2 & ES6D & Uni6Dv2+Geo6D & ES6D+Geo6D  \\ 
%     \midrule
%         Avg  &  59.9 &  82.9 &  91.8 &  92.7 &  88.8 &  91.5 & 93.2 & 91.6 & \textbf{93.6}\\
%     \bottomrule
% \end{tabular}}
% \caption{Evaluation results on the YCB-Video dataset.}
% \label{tab:ycb-ADD}
% \end{center}            
% \end{table*}
\begin{table}[tbp]
\begin{center}
\resizebox{\linewidth}{!}{
\begin{tabular}{l|c|c|c|c|c|c|c}
    \toprule
   & PoseCNN  & PVN3D & FFB6D  & Uni6Dv2 & ES6D & Uni6Dv2+Geo6D & ES6D+Geo6D  \\ 
    \midrule
    Avg  & 59.9 &  91.8 &  92.7 &   91.5 & 93.2 & 91.6 & \textbf{93.6}\\
    \bottomrule
\end{tabular}}
\caption{Evaluation results on the YCB-Video dataset.}
\label{tab:ycb-ADD}
\end{center}
\vspace{-14pt}
\end{table}

\textbf{Evaluation on YCB-Video dataset.}
The quantitative results on the YCB-Video dataset are presented in Tab~\ref{tab:ycb-ADD}. We provide the implementations on the baseline Uni6Dv2 and ES6D, and introducing the Geo6D mechanism brings a 0.1\% and 0.4\% performance gain on the AUC of ADD(S) for full training data. 
Due to the large amount of data in the YCBV dataset, the network can be able to fit some variables, resulting in a relatively small improvement with the full dataset. 
As the result in Tab~\ref{tab:YCB-AUC-es6d}, the fewer training data, our method can achieve greater improvement.

In summary, our Geo6D mechanism makes direct 6D pose estimation methods outperform indirect methods on Occlusion LineMOD and YCB-Video datasets, and achieve comparable accuracy on the LineMOD dataset. 
\begin{table}[t]
\begin{center}
\resizebox{0.99\linewidth}{!}{
\begin{tabular}{l|c|c|c|c|c|c}
    \toprule
 class & PoseCNN   & PVN3D & FFB6D  & Uni6D & Uni6Dv2 & Ours\\
    \midrule
      ape & 9.6  &  33.9 & 47.2  &33.0 & 44.3 & 64.6\\
      can & 45.2  & 88.6 & 85.2  &51.1 & 53.3 & 91.5\\
      cat & 0.9  & 39.1 & 45.7  & 4.6 & 16.7 & 63.2\\
      driller & 41.4 & 78.4 & 81.4  & 58.4 & 63.0 & 82.3\\
      duck & 19.6  & 41.9 & 53.9  & 34.8 & 51.6 & 63.9\\
      \textbf{eggbox} & 22.0  & 80.9 & 70.2  & 1.7 & 4.6 & 95.4\\
      \textbf{glue} & 38.5   & 68.1 & 60.1  & 30.2 & 40.3 & 95.0\\ 
      holepuncher & 22.1  & 74.7 & 85.9  & 32.1 & 50.9 & 82.6\\
       \hline
        Avg & 24.9  & 63.2 & 66.2 & 30.7 & 40.6 & \textbf{79.8} \\ \bottomrule
\end{tabular}}
\caption{Evaluation results on the Occlusion LineMOD dataset. Symmetric objects are denoted in bold. "Ours" is Uni6Dv2+Geo6D.}
\label{tab:ocLineMOD-ACC}
\end{center}
\vspace{-10pt}
\end{table}

\begin{table}[tbp]
\begin{center}
\resizebox{0.99 \linewidth}{!}{
\begin{tabular}{l|c|c|c|c|c|c}
    \toprule
 & PoseCNN & PVN3D & FFB6D & Uni6D & Uni6Dv2 & Ours \\
    \midrule
      ape & 77.0    & 97.3 & 98.4 & 93.7 & 95.7 & 98.3\\
      benchvise & 97.5  &  99.7 & 100.0 & 99.8 & 99.9 & 100.0\\
      camera & 93.5   & 99.6 & 99.9 & 96.0 & 95.8 & 99.6\\
      can & 96.5  & 99.5 & 99.8 & 99.0 & 96.0 & 99.9\\
      cat & 82.1  & 99.8 & 99.9 & 	98.1 & 99.2 & 100.0\\
      driller & 95.0  & 99.8 & 100.0 & 99.1 & 99.2 & 99.8\\
      duck & 77.7   & 97.7 & 98.4 & 90.0 & 92.1 & 97.4\\
      \textbf{eggbox} & 97.1  & 99.8 & 100.0 & 100.0 & 100.0 & 100.0\\
      \textbf{glue} & 99.4  & 100.0 & 100.0 & 99.2 & 99.6 & 100.0\\ 
      holepuncher & 52.8  &  99.9 & 99.8 & 90.2 & 92.0 & 99.7\\
      iron & 98.3   & 99.7 & 99.9 & 99.5 & 98.0 & 100.0\\
      lamp & 97.5   & 99.8 & 99.9 & 99.4 & 98.5 & 99.9\\
      phone & 87.7    & 99.5 & 99.7 &97.4 & 97.7 & 99.8\\
      \hline
      Avg & 88.6   & 99.4 & \textbf{99.7} & 97.0  & 97.2 & 99.6\\   \bottomrule
  \end{tabular}}
  \caption{Results on the LineMOD dataset. Symmetric objects are denoted in bold. "Ours" is Uni6Dv2+Geo6D.}
\label{tab:LineMOD-ADD}
\end{center}
\vspace{-20pt}
\end{table}

\subsection{Ablation study}
In this section, we will analyze the effectiveness of the Geo6D mechanism and compare different reference point generation strategies on Uni6Dv2. All experiments are conducted on the Occlusion LineMOD dataset with 10\% training data.

\begin{table}[t]
\begin{center}
\resizebox{1.0 \linewidth}{!}{
    \begin{tabular}{c|c|c}
    \toprule
    \multicolumn{2}{c|}{Effect of different components in Geo6D mechanism} & \multirow{2}{*}{ADD(S)}  \\
    \cline{1-2} 
    components & setting  \\
    \midrule
    \multirow{2}{*}{(1): $\xyd$  value}  & absolute value &  8.5 \\ 
     & offset value &  \textbf{70.7}	 \\
    \midrule
    \multirow{2}{*}{(2): (1) + Geo head}  & absolute value: $a_i$ &  70.8 \\
     & offset value: $a_i - a_0$ &  71.1	 \\
     & offset value: $\frac{a_i}{d_i} - \frac{a_0}{d_0}$ &  \textbf{72.8}	 \\
    \midrule
    \multirow{2}{*}{(3): (2) + $D \cdot d_0$ input}  & vanilla normalized $\xyd$ with $D \cdot d_0$ &  73.1 \\
     & proposed normalized $\xyd$ with $D \cdot d_0$ &  \textbf{73.7}	 \\
    \midrule
    (4): (3) + $\frac{t_0}{D \cdot d_0}$ input & $\frac{t_0}{D \cdot d_0}$ input & \textbf{74.2}  \\  
    \midrule
    (5): (4) - RGB input & only the camera frame constraints & \textbf{73.7}  \\  
    \bottomrule
    \end{tabular}
}

\caption{Effect of different components in Geo6D mechanism. The best setting in the previous row serves as the benchmark for the next row.}
\label{tab:compoents_ablation}
\end{center}   
\vspace{-10pt}
\end{table}

\begin{table}[t]
\begin{center}
\resizebox{0.85 \linewidth}{!}{
    \begin{tabular}{c|c|c}
    \toprule
    \multicolumn{2}{c|}{Reference point generation strategies} & \multirow{2}{*}{ADD(S)}  \\
    \cline{1-2} 
     $x_0$ and $y_0$ & $d_0$ & \\
    \midrule
    Projection values & nearest depth point &  66.7 \\ 
    Projection values & mean depth point &  72.7	 \\
    mean of visible points & mean of visible points & \textbf{74.2}  \\ 
    \bottomrule
    \end{tabular}
}
\caption{Comparison of generation strategies for the reference point $(x_0, y_0, d_0)$. ``Projection values" denotes calculating from the pin-hole projection equation.}
\label{tab:anchor}
\end{center}   
\vspace{-15pt}
\end{table}

\textbf{Comparison to 3D coordinate normalization.} We introduce the 3D normalization method only converting the coordinate into the offset value to reference points. As shown in Tab~\ref{tab:compoents_ablation} of part (1), the normalized offset value input holds a stable and compact data distribution for the input camera data and achieves 70.7\%. However, it only mentions the camera frame representation breaking the transformation constraints. As shown in Eq~(\ref{equ:translation_equ2}), the translation part is eliminated which is out of optimization. Since the Geo6D endues the data representation with both camera and object constraints, the performance improves to 74.2\%.

\textbf{Effect of different components of Geo6D mechanism}
To analyze the effect of each input and output components adjustment, an ablation study is conducted on different components in the proposed geometric constraints equation Eq~(\ref{equ:new_prj}).
As reported in Tab~\ref{tab:compoents_ablation}, the $\Delta \xyd$ significantly improves by solving the distribution gap issue. Based on the offset representation, introducing the object frame constraints by Geo head shows an advanced performance when choosing the optimization target following the proposed relative offset $\Delta ABC$ as shown in Eq~(\ref{equ:new_prj}). Taking $a$ channel as an example, the relative offset format $\frac{a_i}{d_i} - \frac{a_0}{d_0}$ follows the equation and achieves the better performance gain.
Compared with only applying the Geo head, additional input of the components $D \cdot d_0$ and $\frac{t_0}{D \cdot d_0}$ as the camera frame constraints can indeed advance the performance.
Furthermore, as the result in the $(3)$ row, the proposed relative offset representation ($\frac{x_i}{d_i} - \frac{x_0}{d_0}$) outperforms the vanilla normalization ($x_i - x_0$), demonstrating the advantage of introducing the translate vector into optimization. 
The last row is performed without the RGB input to show that the proposed Geo6D mechanism can perform well without visible feature-pose constraints.  

\begin{figure}[t]
    \centering
    \includegraphics[width=0.8\linewidth]{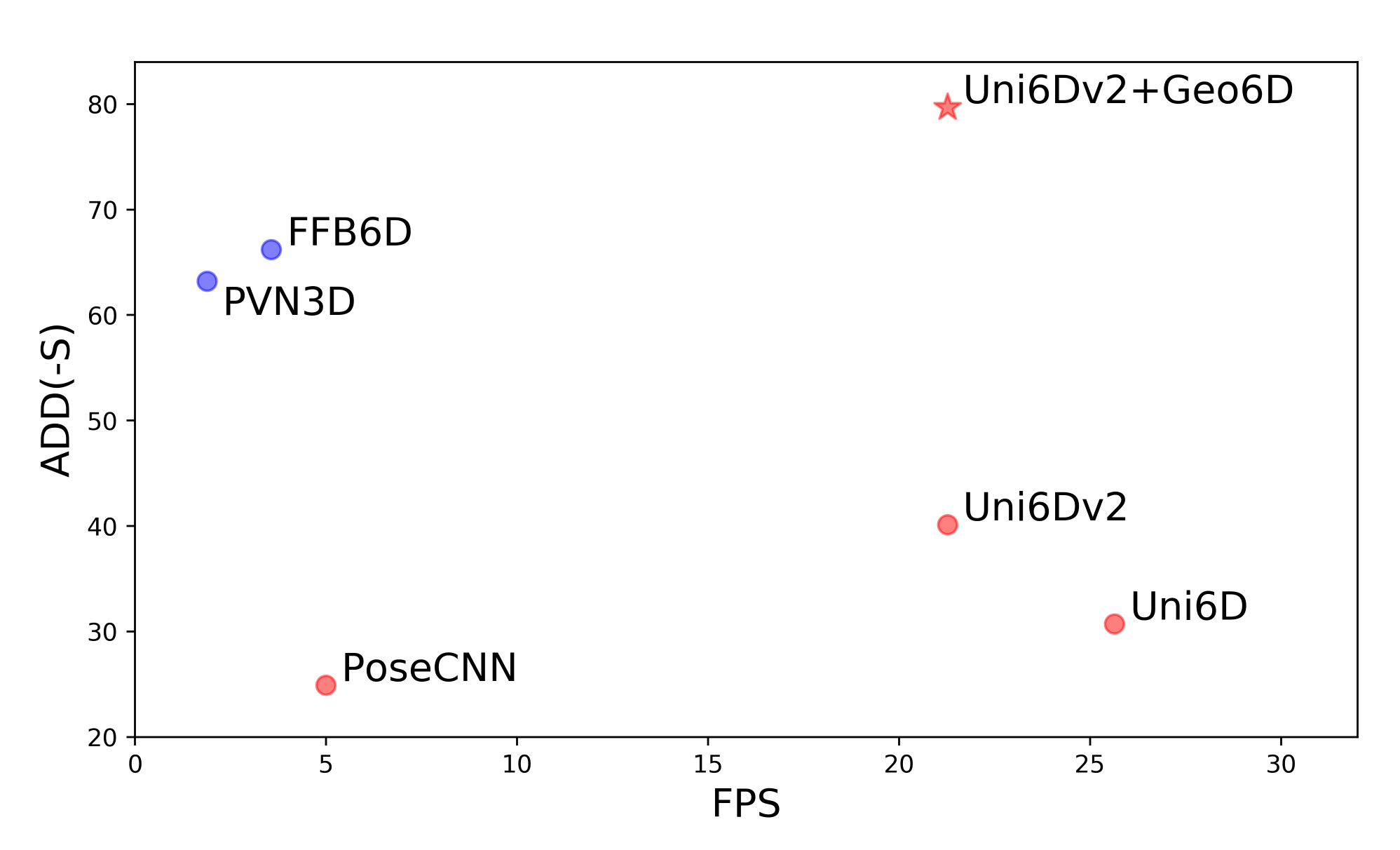}
    \vspace{-5pt}
    \caption{Comparison to speed and accuracy on the Occlusion LineMOD with full amount training data: Uni6Dv2+Geo6D achieves state-of-the-art accuracy in both direct regression methods (\textcolor{red}{red dots}) and indirect regression methods (\textcolor{blue}{blue dots}) methods.}
    \label{fig:performance}
    \vspace{-10pt}
\end{figure}

\textbf{Comparison to speed and accuracy.}
% We analyze the effect of different relative offset representation strategies, including the vanilla 3D normalization and the proposed relative offset representation. 
%
% When only introducing the vanilla normalization into UniDv2 to regress a relative offset, the performance of Uni6Dv2 improves by 27.2\%, achieving 67.8\% w.r.t ACC-ADD(S).
%
% When adopting the proposed Geo6D mechanism to Uni6Dv2, there is an obvious performance gain from 40.6\% to 79.8\%, and it shows a 12\% advanced improvement compared with Uni6Dv2 equipped with 3D vanilla normlization.
%
When adopting the proposed Geo6D mechanism to Uni6Dv2, as shown in Fig~\ref{fig:performance}, it merely does not introduce extra time consumption and helps Uni6Dv2 outperform the current SOTA RGB-D method FFB6D~\cite{ffb6d} 13.6\% while 5.6 times faster than it.
% Moreover, for analyzing efficiency, as shown in Fig~\ref{fig:performance}, the relative offset representation merely does not introduce extra time consumption and helps Uni6Dv2 outperform the current SOTA RGB-D method FFB6D~\cite{ffb6d} 13.6\% while 5.6 times faster than it.
%
The improvement demonstrates that the proposed geometric constraints can facilitate direct network higher performance without scarifying the efficiency.

\textbf{Comparison to different reference point generation strategies.}
To build a stable reference point, we attempt three different reference point generation strategies, as shown in Tab~\ref{tab:anchor}. 
The first strategy selects the 2D coordinate UV value $(u_{0}, v_{0})$ of the ROI region center, the nearest point's depth as reference point depth $d_0$, and then calculates $x_{0}$ and $y_{0}$ based on pin-hole projection equation. 
The ROI center as the reference point in the 2D image plane maintains a stable UV encoding and attempts to use the center point's depth to project the reference point in the 3D space.
However, since the center point in ROI is likely in the outside or margin of the object due to the occlusion or the irregular, the depth value may be lost or a noise point from the sensor.
To alleviate this problem, the second strategy adopts the mean depth value of the object region as the reference depth, which improves the stability of the Geo6D mechanism, achieving a 6\% improvement from 66.7\% to 72.7\%. 
Besides the reference point selection based on the UV, the third strategy adopts the mean XYD of visible points as the reference point.
Compared with the selection from UV, it is prone to consider the 3D shape and shows higher performance.
Hence, whatever the reference point generation considers the 2D shape or 3D shape, the Geo6D mechanism can demonstrate its effectiveness.
% \subsection{Visualization}
% Visualization results of the Occlusion LineMOD dataset are shown in Fig~\ref{fig:vis}. We can observe that Uni6Dv2+Geo6D could predict a more precise and robust 6D pose on the heavily occluded objects compared to other methods. Moreover, benefiting from the Geo6D mechanism, the orientation of objects with only a few visible points is correctly predicted, which proves that the proposed Geo6D mechanism improves the robustness and generalization of the network.

% \begin{figure}[tbp]
%     \centering
%     \includegraphics[width=0.49\textwidth]{figures/vis.png}
%     \caption{Qualitative results of 6D pose estimation on the Occlusion LineMOD dataset. 3D norm denotes that only adopts the vanilla normalization.}
%     \label{fig:vis}
% \end{figure}
\textbf{Comparison to different segmentation results} Most of these methods
including our two baseline Uni6Dv2 and ES6D adopt the same segmentor MaskRCNN. All experiments hold the same upstream segmentation result with their baselines. To evaluate the effect of Geo6D in different segmentation results, we conduct an experiment on the LM-O dataset to adopt
the Ground Truth (GT) mask instead of the predicted mask, where the GT mask shows a slight 0.3\% improvement from 79.8\% to 80.1\%. It shows the proposed Geo6D based on the predicted segmentation result reaching nearly the upper bound performance and demonstrates the robustness of the segmentation results.

\textbf{Comparison to scaling different parts of ADD Loss.} The imbalance of the translation and rotation part in the ADD loss is due to the issue of unlimited output space of the translation part. Scaling the two parts into a comparable level leads to the challenging translation part without enough optimization. We conduct an experiment based on the analysis in Fig~\ref{fig:balanced_loss}, increasing rotation part weight 4 times, where the performance decreased from 40.6\% to 15.4\%. It supports that only scaling the two-part loss can not achieve the same effectiveness as our balanced ADD loss.

\section{Conclusion}

In this paper, we propose a novel geometric constraints learning approach for 6D pose estimation that achieves state-of-the-art performance on multiple benchmarks.  First, a non-ill-conditioned 6D pose transformation is derived. Geo6D rebuilds the input data based on the derived transformation and employs a Geo head to strengthen the point-to-point relationship as constraints between the camera frame and the object frame in a leaning-friendly manner. Extensive experiments show that Geo6D greatly simplifies the task of 6D pose estimation and can be plugged into various direct methods such as Uni6Dv2 and ES6D. Experiments also indicate that Geo6D achieves higher gains with less training data. We believe the Geo6D mechanism has the potential to inspire other 3D tasks, such as category-level pose estimation and 3D object detection using RGB-D or Lidar data. 

In terms of limitations, Geo6D's current reference point generation strategy requires that the mean point be close to the centroid of the majority of objects. Geo6D struggles to generate a stable reference point on the CAD model when the mean point of irregular objects is far from the centroid. When dealing with irregular objects, more effective generation strategies must be investigated.
\bibliography{ref}

\end{document}